\documentclass[a4paper, 10pt, conference]{ieeeconf}      %
\usepackage{FG2027}
\usepackage{hyperref}
\usepackage{graphicx}
\usepackage{amsmath}
\usepackage{capt-of}   %
\usepackage{booktabs}
\usepackage{amssymb} %
\FGfinalcopy %

\IEEEoverridecommandlockouts                              %
\def\FGPaperID{212} %

\title{\LARGE \bf
DuoGesture: Motion-Grounded Semantic Conditioning and Biomechanical Beat Priors for Co-Speech Gesture Generation}

\usepackage{wrapfig} %
{\ignorespaces}
{\ignorespaces}
{\ignorespaces}

\usepackage{multirow}
\usepackage[table]{xcolor}
\usepackage[normalem]{ulem}
\definecolor{rankbest}{HTML}{C6EFCE}
\definecolor{ranksecond}{HTML}{D9EAF7}
\definecolor{rankthird}{HTML}{FFF2CC}

\newcommand{\bestcell}[1]{\cellcolor{blue!20}\textbf{#1}}
\newcommand{\secondcell}[1]{\cellcolor{blue!10}#1}
\newcommand{\thirdcell}[1]{\cellcolor{blue!5}#1}

\definecolor{semcolor}{RGB}{52, 120, 198}    %
\definecolor{beatcolor}{RGB}{230, 175, 45}   %
\definecolor{mgsccolor}{RGB}{40, 140, 110}   %
\definecolor{ibpcolor}{RGB}{180, 50, 80}     %

\author{\parbox{16cm}{\centering
    {\large Ferdinand Paar$^{1,2,*}$, Lanmiao Liu$^{1,3,*}$, Asl{\i} {\"O}zy{\"u}rek$^{1,2}$,
     Serge Thill$^{2}$ and Esam Ghaleb$^{1}$}\\[0.5ex]
    {\normalsize
    $^{1}$Max Planck Institute for Psycholinguistics, Nijmegen, The Netherlands\\
    $^{2}$Radboud University, Nijmegen, The Netherlands\\
    $^{3}$Utrecht University, Utrecht, The Netherlands\\[0.5ex]
    {\tt\small \{ferdinand.paar, asli.ozyurek, esam.ghaleb\}@mpi.nl,
     l.liu03@uu.nl, serge.thill@donders.ru.nl}}}%
    \thanks{$^{*}$Equal contribution.}%
    \thanks{Preprint. Under review.}%
}

\begin{document}

\ifFGfinal
\thispagestyle{empty}
\pagestyle{empty}
\else
\author{Anonymous FG2027 submission\\ Paper ID \FGPaperID \\}
\pagestyle{plain}
\fi
\maketitle
\begin{abstract}
Co-speech gesture generation requires both semantic expressivity and biomechanically plausible rhythmic motion. Existing holistic gesture models mix lexically grounded semantic gestures with frequent prosody-aligned beat gestures. This limits semantic grounding, speech-motion alignment, and kinematic smoothness. We propose \emph{DuoGesture}, a neuro-inspired and biomechanically informed approach that decomposes co-speech gesture synthesis into semantic and beat streams. The two streams are coordinated by a \emph{Semantic Variational Information Bottleneck}, a stochastic frame-level gate that learns when semantic gestures should override rhythmic beat motion.
The semantic stream is controlled by \emph{Motion-Grounded Semantic Conditioning}, which replaces purely linguistic word embeddings with motion-language representations to provide motion-aligned semantic priors for long-tailed lexical triggers of gestures. The beat stream is further regularised by an \emph{Inertial Beat Prior}, an anthropometry-weighted arm-chain module that reduces jitter and improves rhythmic consistency without constraining semantic frames. Objective evaluations and subjective experiments show that DuoGesture outperforms strong baselines, while component ablations confirm the complementary roles of semantic grounding, stochastic stream selection, and biomechanical regularisation. We encourage readers to view our demos and qualitative illustrations, which are available on the anonymized project page:
\href{https://duogesture.github.io/DuoGesture/}{https://duogesture.github.io/DuoGesture/}.
\end{abstract}

\section{Introduction}\label{sect:intro}
Speech and co-speech gestures form an integrated communicative system, yet gesture types play different roles and are processed accordingly. Reliable co-speech gesture generation is therefore important for embodied agents, virtual avatars, and accessible speech-driven animation systems, where gestures affect perceived naturalness, communicative clarity, and user trust. Beat gestures align with prosody and rhythm, supporting turn-taking and the flow of interaction, while semantic gestures (deictic, iconic, metaphoric) are related to lexical content and occur more sparsely~ \mbox{\cite{mcneill1992hand}}. Cognitive neuroscience further indicates that manual gesture comprehension recruits partially dissociable but interacting networks: parietal--premotor circuits associated with visuomotor structure, and inferior-frontal/temporal circuits implicated in communicative meaning and gesture--speech integration \mbox{\cite{tipper2015body, arbib2012brain}}. We adopt this view as a computational stance, where co-speech gesture should be modelled as coupled streams rather than as a single motion process, where recent work has pointed out the benefit of this dual stream approach \cite{liang2022seeg, zhang2025SemTalk}. However, current gesture generation still has \emph{three limitations}: (i) misalignment between gesture and speech timing, (ii) poor semantic expressivity, and (iii) jittery beat motion that lacks biomechanical principles.

We argue that these limitations stem from two main modelling drawbacks. First, semantic conditioning is usually derived from linguistic embeddings derived from off-the-shelf text or vision-language encoders such as BERT~\cite{devlin2019bert} or CLIP~\cite{radford2021learning}, whose representations are learned from linguistic or image-text supervision rather than from body-motion dynamics. 
This leaves a \emph{linguistic-motion gap}, where embeddings encode \emph{what was said} rather than \emph{what motion should accompany it}. This is especially problematic for long-tailed semantic triggers that appear rarely, or never, during training. Second, beat gestures are learned without physical or biomechanical constraints. These choices create gaps: a \emph{semantic gap}, where text embeddings provide weak priors for what a gesture should look like, and a \emph{beat gap}, where rhythmically aligned gestures may achieve reasonable average objective metrics but remain prone to over-smoothing and acceleration artefacts.

\begin{figure*}
    \centering
    \includegraphics[width=0.9\linewidth]{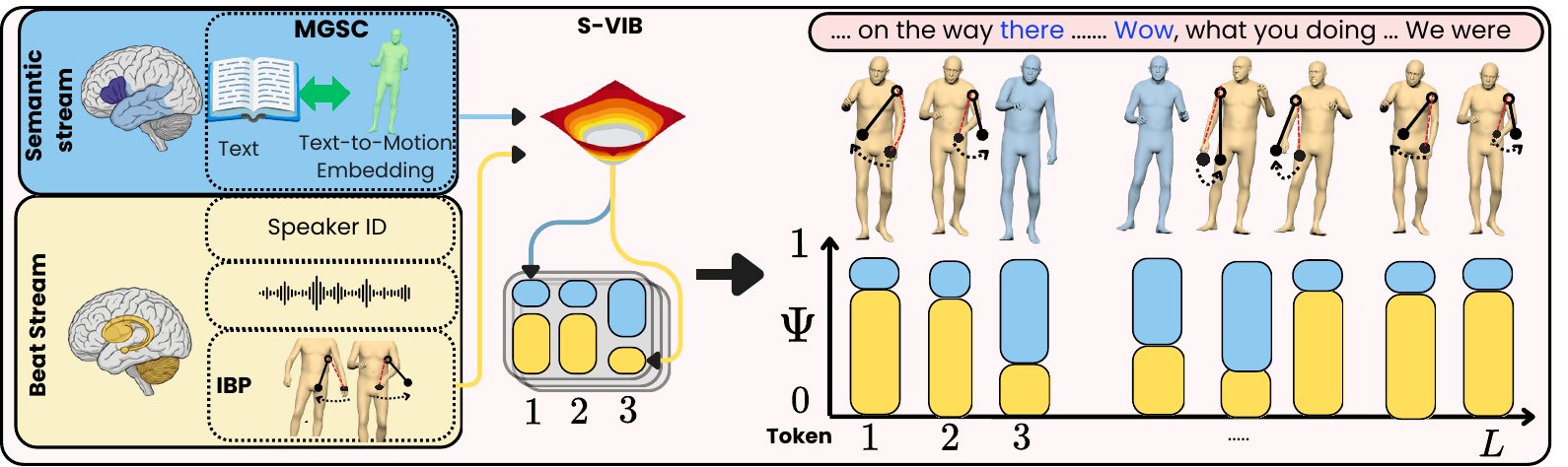}
    \caption{DuoGesture integrates 
    \textcolor{teal}{\textit{motion-grounded semantic conditioning}} and an 
    \textcolor{purple}{\textit{inertial prior}} to ensure semantic 
    expressivity and rhythmic smoothness.}
    \label{fig:main_figure}
\end{figure*}

To overcome these limitations and gaps, we propose \emph{DuoGesture}, a model shown in Figure~\ref{fig:main_figure}, that integrates lexically grounded semantic motion from prosody-aligned beat motion while allowing them to interact. Supported by our auxiliary analysis in Supplementary Materials Section 1, which shows distinct temporal and kinematic profiles for semantic- and beat gestures, DuoGesture uses stream-specific conditioning and regularisation. The semantic stream is conditioned by \emph{Motion-Grounded Semantic Conditioning} (MGSC), which leverages a pretrained text-to-motion representation to provide motion-aligned cues for long-tailed lexical triggers of gestures. The beat stream is regularised by an \emph{Inertial Beat Prior} (IBP), an anthropometry-weighted arm-chain loss that encourages smooth rhythmic motion while remaining inactive on semantic frames. The streams are weighted by a \emph{Semantic Variational Information Bottleneck} (S-VIB), which learns when semantic motion should override beat motion. Our novel contributions are summarised as follows:

\begin{itemize}
    \item We introduce \textbf{MGSC}, a motion-grounded semantic conditioning module that, to our knowledge, is the first to drive a co-speech gesture generator from a
    motion-aligned text representation rather than a purely linguistic or image-text
    one. 
    
    \item We design \textbf{IBP}, an anthropometry-weighted arm-chain regulariser that reduces beat-motion jitter and improves rhythmic consistency without constraining semantic frames.
    
    \item We present \textbf{S-VIB} that learns when semantic gestures should override beat motion using a stochastic frame-level gate.
    
    \item On the BEAT2 benchmark, DuoGesture improves distributional realism as measured by Fr\'echet Gesture Distance (FGD). We further report two metrics that test our contributions directly: a semantic motion score on semantically annotated frames, and a measure of motion smoothness. DuoGesture leads on all three, and a user study rates it above competitive baselines.
\end{itemize}

\section{Related Work}\label{sect:related_work}

\paragraph{Co-speech Gesture Generation.} Current work has been shaped by two-stage hierarchical quantisation paradigms. Early architectures, e.g., by TalkSHOW~\cite{yi2023generating} and ProbTalk~\cite{liu2024towards}, established VAE-based stochastic priors and category-contingent decoding for multifaceted body dynamics. This trajectory toward granularity was extended by EMAGE~\cite{liu2024emage}, which proposed a spatially-decoupled tokenisation scheme across facial, manual, and corporal streams.
PyraMotion~\cite{yinpyramotion} leveraged a multi-resolution Anchor-based Pyramid VQ-VAE to minimise distributional divergence, yielding a competitive FGD. 

The aforementioned models assume \textbf{kinematic
homogeneity} of co-speech gestures, treating it as a singular process. However, the difference between beat and semantic gestures is foundational in behavioural and cognitive sciences \cite{tipper2015body, arbib2012brain}, and researchers have attempted to exploit it. SEEG~\cite{liang2022seeg}
encodes lexical and hand-crafted onset cues in two branches, but merges them in a
single shared upper-body decoder. LivelySpeaker~\cite{zhi2023livelyspeaker} used a sequential two-stage pipeline where its semantic-aware generator maps a whole-script CLIP embedding to motion, after which a separately trained audio-conditioned diffusion model adds noise and rhythmically refines that motion. Closest to ours, SemTalk~\cite{zhang2025SemTalk} combines
the streams via a deterministic frame-level gate. However, in our analysis, this gate tends to favour the
dense beat pathway, and its semantic branch remains bottlenecked by shallow lexical conditioning
and an absence of physical regularisation.

DuoGesture exploits the theoretical differences between beat and semantic gestures through how the two streams are conditioned, selected, fused, and regularised.
Motion-Grounded Semantic Conditioning (MGSC) exploits time-aligned
features from a pretrained text-to-motion space. A Semantic Variational Information Bottleneck
(S-VIB) uses a stochastic gate so the two predictions are selectively combined rather than uniformly aggregated. Finally, an
Inertial Beat Prior (IBP) imposes anthropometrically weighted regularisation on the proximal arm
chain at beat frames only.

\paragraph{Cross-Modal Semantic Grounding Gesture Generation.}
A major gap in co-speech gesture generation is the \textit{linguistic-kinematic gap}, where word embeddings (e.g., BERT \cite{devlin2019bert}) are insufficient to encapsulate the morphology of human motion. To solve this, Semantic Gesticulator~\cite{zhang2024semantic} leverages LLM-based lexical parsing, yet the resulting representations remain inherently decoupled from the physical motion manifold. While GestureDiffuCLIP~\cite{ao2023gesturediffuclip} employs CLIP latents to enforce semantic consistency, it operates within a text-centric latent space, often overlooking the fine-grained temporal dynamics of gesticulation. The RAG-Gesture~\cite{mughal2025retrieving} introduced non-parametric synthesis by retrieving explicit motion exemplars at inference time. Our work resolves this; instead of performing an explicit search, we \textit{exploit} the structured knowledge of a motion-aligned encoder (TMR~\cite{petrovich2023tmr}) into our generative pipeline.

\paragraph{Physically-Consistent Motion Synthesis.}
While physics-based constraints have been explored in character animation via differentiable simulators or post-hoc projections~\mbox{\cite{peng2018deepmimic, yuan2023physdiff}}, their integration as inductive biases within gesture models remains under-explored. In this work, we diverge from computationally expensive simulation-based approaches by incorporating \textit{anthropometric priors} directly into the generator's rhythmic stream. Specifically, we leverage De Leva's segment mass distributions~\mbox{\cite{de1996adjustments}} to derive joint-specific exponential moving average time constants.
This formulation provides a computationally efficient, training-time-only regularisation that imposes no inference overhead. Concretely, IBP uses an auxiliary loss on arm-chain motion, with the De Leva weights setting each joint's contribution; unlike full physics simulation, it leaves the inference pipeline untouched.

\section{Method}
\label{sec:methodology}
DuoGesture is a two-stage latent generator. Stage~1 (Sec.~\ref{sec:formulation}) is a
region-wise RVQ-VAE tokenizer. Our contributions are in Stage~2:
(i) \emph{Motion-Grounded Semantic Conditioning} (MGSC, Sec.~\ref{sec:MGSC})
grounds semantic conditioning in a text-to-motion latent space; (ii) \emph{Inertial Beat Prior} (IBP, Sec.~\ref{sec:ibp}) regularizes velocity consistency along the proximal arm chain at training time, injecting an anthropometric inductive bias into the beat stream and (iii) \emph{Semantic Variational
Information Bottleneck} (Sec.~\ref{sec:svib}) predicts per-frame weights
$\Psi$ under an information-bottleneck constraint.
\subsection{Problem Formulation}
\label{sec:formulation}
A motion sequence of length $L$ is split into four body regions
$\mathcal{R}=\{\mathrm{hand},\mathrm{upper},\mathrm{lower},\mathrm{face}\}$,
$\mathbf{G}=\{\mathbf{G}^{r}\}_{r\in\mathcal{R}}$ with
$\mathbf{G}^{r}\in\mathbb{R}^{L\times J_{r}}$. 
Stage~2 conditions on
HuBERT audio features $e_a\in\mathbb{R}^{L\times 1024}$, a speaker identity
embedding $\mathrm{ID}$, motion-grounded semantic features
$\mathbf{S}^m\in\mathbb{R}^{L\times 256}$ the per-frame output of the MGSC module (Sec.~\ref{sec:MGSC}) and
a 4-frame seed pose $\tilde{\mathbf{p}}$. The target is the discrete
latent code $\mathcal{Z}^{q}=\{\mathbf{Z}^{q}_{r}\}_{r\in\mathcal{R}}$
produced by the Stage~1 quantiser on the ground-truth motion. 

\paragraph{Stage 1: Regional RVQ-VAE Tokeniser.}
\label{sec:stage1}
We adopt the regional RVQ-VAE of~\cite{liu2024emage, zhang2025SemTalk, liu2025semges,liu2026semconflow}: one encoder--quantiser--decoder triple
$(\mathcal{E}^{r},\mathcal{Q}^{r},\mathcal{D}^{r})$ per region, trained with the standard reconstruction and codebook
commitment losses. The stage-1 components are kept frozen once moving to Stage 2, decoupling (Stage~1) from temporal motion synthesis (Stage~2), and enabling the generator to operate in a structured,
low-dimensional latent space.

\subsection{DuoGesture}
\label{sec:stage2}

DuoGesture separates beat and semantic motion and adaptively fuses them via a dedicated module. The \textit{beat backbone} $f_b$ produces per-region latents from audio, speaker identity, and seed pose alone:
\begin{equation}
    f_b:\bigl(e_a,\;\mathrm{ID},\;\tilde{\mathbf{p}};\;\theta_b\bigr)
    \;\longrightarrow\; Z_r^{\mathrm{b}}, \quad r\in\mathcal{R},
    \label{eq:beat}
\end{equation}
where $Z_r^{\mathrm{b}}$ is the beat codebook per region. This beat branch is modulated by a semantic branch via a dedicated weighting module, namely \textit{S-VIB}. First, the \textit{MGSC} module assembles per-frame lexical ($e_s$), motion-style ($e_m$), and emotion ($e_\varepsilon$) embeddings into a motion-grounded semantic feature, and \textit{S-VIB} produces per-frame the following:

\begin{equation}
f_{s\text{-}vib}:
\bigl(e_s,e_m,e_\varepsilon,e_a;\theta_{s\text{-}vib}\bigr)
\longrightarrow
\bigl(\mathbf{S}^m,\Psi\bigr).
\label{eq:semantic_module}
\end{equation}

Where $\mathbf{S}^m\in\mathbb{R}^{L\times256}$ and
$\Psi\in[0,1]$. We then use the semantics features $\mathbf{S}^m$ to impose semantics on beat frames' output through the semantic branch: 
\begin{equation}
    f_s:\bigl(e_a, \;\mathrm{ID}, \Psi, \;\mathbf{S}^m;\;\theta_s\bigr)
    \;\longrightarrow\; Z_r^{\mathrm{s}}, \quad r\in\mathcal{R};
    \label{eq:semantic}
\end{equation}

where the output of the semantic branch $Z_r^{\mathrm{s}}$ is then injected into $Z_r^{\mathrm{b}}$ via a dedicated module to produce the fused codebook $Z_r$ = $ f_{fusion}\bigl(Z_r^{\mathrm{b}},\; Z_r^{\mathrm{s}},  \Psi; \theta_{fusion})$.
The resulting codebook is then fed into the Stage-1 decoder $\mathcal{D}^r$, which maps the result to joint space: $\hat{\mathbf{G}}^r = \mathcal{D}^r\!\bigl(Z_r)$.

\begin{figure*}
    \centering
    \includegraphics[width=0.9\linewidth]{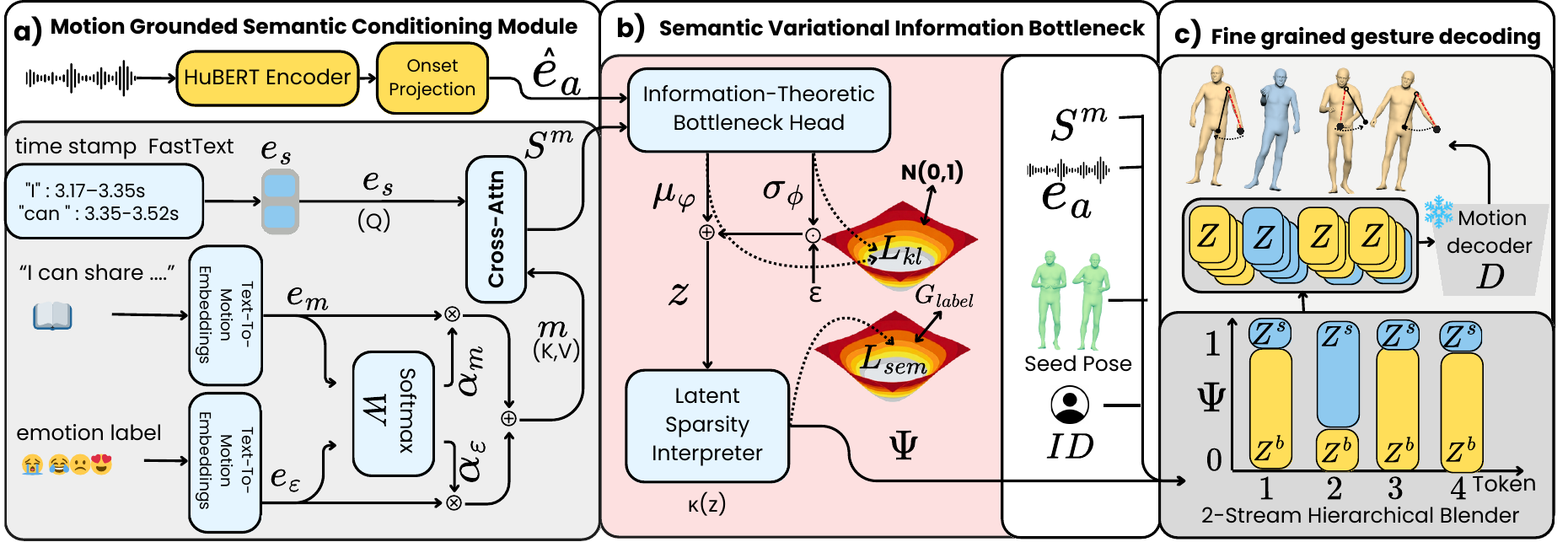}
    \caption{DuoGesture pipeline.  (a) MGSC fuses lexical semantics, motion-style, and emotion embeddings through cross-attention to produce the motion-grounded semantic representation $\mathbf{S}^m$.  (b) S-VIB combines $\mathbf{S}^m$ with the HuBERT timing projection $\hat{e}_{a}$ to infer when semantic gestures should be activated and what semantic content they should express. The bottleneck samples $\mathbf{z}\tau \sim q\phi(\mathbf{z}\tau)$ under $\mathcal{L}{\mathrm{kl}}$ regularisation, and maps it to the semantic gate $\Psi$ with semantic supervision. (c) Fine-grained decoding blends beat codebooks $Z_r^{\mathrm{b}}$ (yellow) and semantic codebooks $Z_r^{\mathrm{s}}$ (blue) using $\Psi$.}
    \label{fig:pipeline}
\end{figure*}

\subsubsection{Motion-Grounded Semantic Conditioning}
\label{sec:MGSC}
To bridge the gap between abstract semantic representations and their
corresponding kinematic realisations, we propose the MGSC module. Existing gesture-generation methods typically use
off-the-shelf text or vision-language encoders, for example
FastText~\cite{bojanowski2017enriching}, BERT~\cite{devlin2019bert}, or
CLIP~\cite{radford2021learning}, whose representations are learned from
linguistic or image-text supervision rather than from body-motion dynamics.
In contrast, MGSC introduces a motion-grounded conditioning pathway that aligns semantic cues with their kinetic realisations. As illustrated in Fig.~\ref{fig:pipeline}(a), MGSC produces a per-frame semantic feature $\mathbf{S}^m \in \mathbb{R}^{L \times 256}$ that conditions the semantic generator $g_{\mathrm{sem}}$ and feeds S-VIB. It assembles three streams, all projected to 256 dimensions: (i)~$e_s \in \mathbb{R}^{256}$, a per-frame FastText~\cite{bojanowski2017enriching} embedding of the word spoken at frame $l$, from BEAT2's forced-alignment timestamps (hence it gives word timing); (ii)~$e_m$, an utterance-level motion-style embedding from Text-To-Motion~\cite{petrovich2023tmr}; and (iii)~$e_\varepsilon$, an emotion embedding from Text-To-Motion, using emotion labels provided by BEAT2~\cite{liu2022beat}. $e_m$ and $e_\varepsilon$ are blended by a learned softmax gate into a fused memory $\mathbf{m}$:
\begin{equation}
    \boldsymbol\alpha = \mathrm{softmax}\left( \mathbf{W}_\alpha [e_m; e_\varepsilon] \right), \quad \mathbf{m} = \alpha_{(m)} e_m + \alpha_{(\varepsilon)} e_\varepsilon.
\end{equation}
$e_s$ then queries this memory via cross-attention to produce the final semantic feature: $
    \mathbf{S}^m = \mathrm{MLP}\left( \mathrm{CrossAttn}\left( Q=e_s,\; K\!=\!V\!=\!\mathbf{m} \right) \right) \in \mathbb{R}^{L \times 256}$.

\subsubsection{Semantic Variational Information Bottleneck}
\label{sec:svib}
We want a gate to select a semantic gesture at the right moment. For this, the gate needs to consider what and when to gesture semantically. At the same time, this bottleneck (gate) should not collapse into a single path: either all semantic or all beat. For this reason, we propose S-VIB. As illustrated in Fig.~\ref{fig:pipeline}(b), S-VIB operates on two streams: the MGSC output $\mathbf{S}_m\in \mathbb{R}^{256}$ (\emph{what} to gesture, derived from $e_s$) and a low-capacity HuBERT timing projection $\hat{e}_{a} \in \mathbb{R}^{64}$ (\emph{when} to gesture). $\hat{e}_{a}$ is obtained by passing HuBERT features $e_a$ through a convolutional encoder and bottlenecking to 64 dimensions; it is entirely independent of the FastText word embedding $e_s$.

The \textbf{Information-Theoretic Bottleneck Head} maps $\mathbf{S}^m$ and $\hat{e}_{a}$ to two 16-dimensional outputs via separate linear heads: a mean $\boldsymbol\mu_\phi \in \mathbb{R}^{16}$ and a log-variance $\log\boldsymbol\sigma^2_\phi \in \mathbb{R}^{16}$. A stochastic sample is drawn via the reparametrisation trick, $\mathbf{z} = \boldsymbol\mu_\phi + \exp\!\bigl(\tfrac{1}{2}\log\boldsymbol\sigma^2_\phi\bigr) \odot \boldsymbol\epsilon$ with $\boldsymbol\epsilon\sim\mathcal{N}(0,I)$, and passed to the \textbf{Latent Sparsity Interpreter} $\kappa$ (two-layer MLP, $\kappa:\mathbb{R}^{16}\to\mathbb{R}^2$), which outputs two-dimensional beat/semantic logits; the semantic probability is the per-frame gate: $
    \Psi = \mathrm{softmax}\bigl( \kappa(\mathbf{z}) \bigr)_{\mathrm{sem}} \in [0, 1]$.
The outputs $\boldsymbol\mu_\phi$ and $\log\boldsymbol\sigma^2_\phi$ are regularised against the standard Gaussian prior $\mathcal{N}(0,I)$ via a KL divergence~\cite{alemi2017deep, kingma2016improved}:

\begin{equation}
\begin{aligned}
\mathcal{L}_{\mathrm{kl}}
&=
D_{\mathrm{KL}}\!\left(
\mathcal{N}(\boldsymbol\mu_\phi,\boldsymbol\sigma_\phi^2)
\,\middle\|\,
\mathcal{N}(0,I)
\right) \\
&=\frac{1}{2}\sum_{d=1}^{Z}
\Bigl[
\mu_{\phi,d}^{2}
+ e^{\log\sigma_{\phi,d}^{2}}
- \log\sigma_{\phi,d}^{2}- 1\Bigr].
\end{aligned}
\end{equation}
In practice, we apply a per-dimension free-bits floor~\cite{kingma2016improved} ($\lambda_{\mathrm{fb}}=0.50$ nats, $Z=16$), whose per-dim KL falls below $\lambda_{\mathrm{fb}}$ receive no gradient, preventing the optimiser from over-regularising uninformative dimensions while preserving capacity for those that carry semantic signal.
The gate $\Psi$ is trained to predict the per-frame semantic annotations $s_\tau \in \{0,1\}$ via a semantic loss $\mathcal{L}_{sem}$.

\subsubsection{Hierarchical Blender}
\label{sec:hierarchical_refiner}
\begin{figure*}[t!]
    \centering
    \includegraphics[width=.91\linewidth]{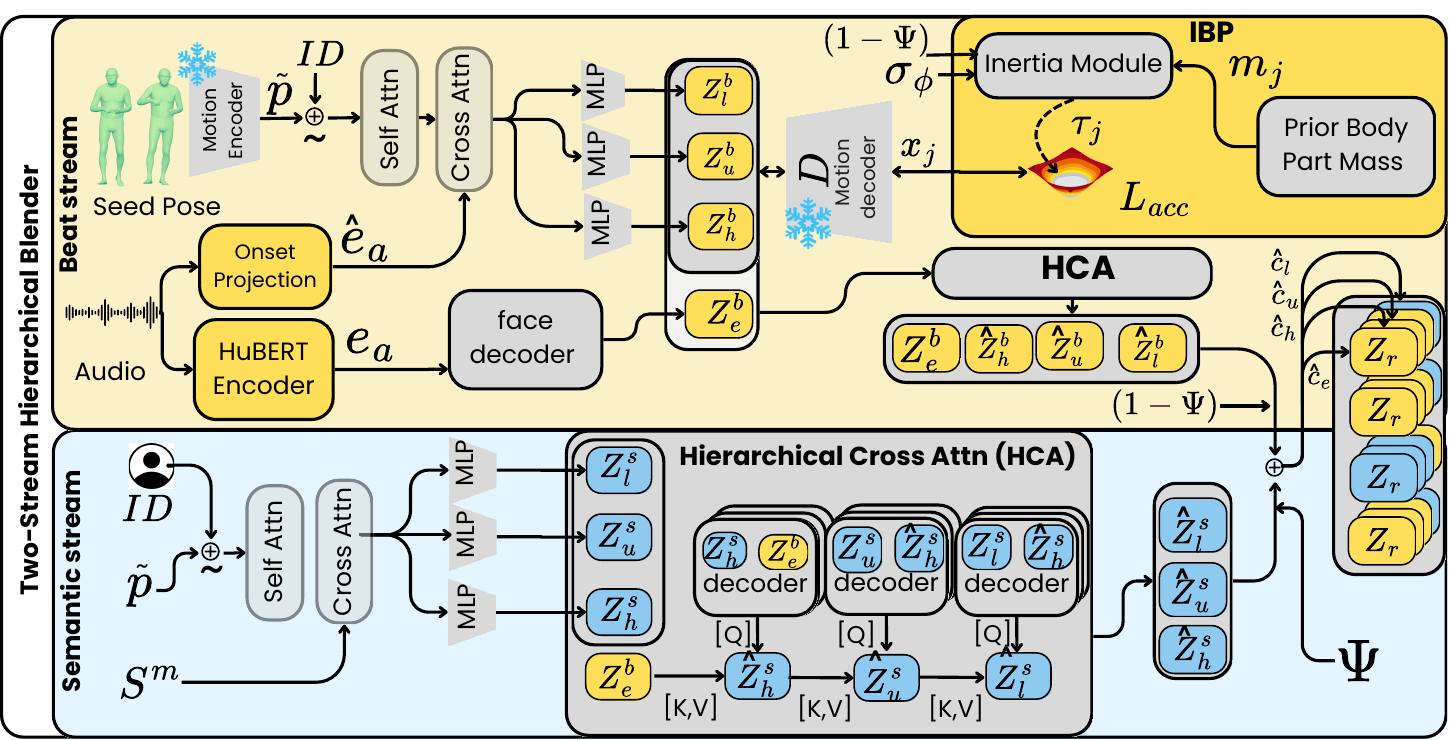}
    \caption{Two-Stream Hierarchical Blender. The beat stream encodes the seed pose $\tilde{\mathbf{p}}$ and speaker ID to predict region-wise beat latents, with the face handled by a separate audio-conditioned decoder. Arm-chain beat latents are decoded during training and regularised by IBP through a $\tau_j$-weighted smoothness loss. The semantic stream conditions parallel region-wise latents on the gated semantic feature. Hierarchical cross-attention refines both streams across body regions.}
    \label{fig:hierarchical_refiner}
\end{figure*}

The hierarchical blender (Fig.~\ref{fig:hierarchical_refiner}) produces per-region beat residuals $Z_r^{\mathrm{b}}$ and semantic residuals $Z_r^{\mathrm{s}}$, decoded by the frozen Stage-1 decoders. It has two parallel branches that differ in their conditioning and regularisation, reflecting the asymmetry between beat and semantic gesture production.
\textbf{Beat stream}, (top, yellow in Fig.~\ref{fig:hierarchical_refiner}) takes as input the masked seed pose $\tilde{\mathbf{p}}$ embedding $\in\mathbb{R}^{T\times 768}$, where we add speaker identity $\mathrm{ID}$ and periodic positional encodings are applied before a self-attention pass. Speech onset projection ( $\hat{e}_a$) is integrated with the resulting embeddings through a cross-attention layer. Three region-specific MLPs then project the refined beat region latents $Z_u^{\mathrm{b}}$, $Z_l^{\mathrm{b}}$, $Z_h^{\mathrm{b}}$. The face latent $Z_e^{\mathrm{b}}$ is obtained from a separate face decoder conditioned on the HuBERT audio feature $e_a$.

\paragraph{Inertial Beat Prior (IBP).}\label{sec:ibp}
On beat stream, we apply our proposal to constrain the upper-body and hand movements with biomechanical constraints. For this reason, the beat latents $Z_u^{\mathrm{b}}$, $Z_l^{\mathrm{b}}$, $Z_h^{\mathrm{b}}$ are decoded by the frozen Stage-1 decoder into raw (rot6d) poses $\mathbf{x}_{j,t}$ and an \textbf{Inertia Module} (fed by body-part masses $m_j$, gate $(1\!-\!\Psi)$, and VIB posterior variance $\sigma^2_{\phi}$) produces per-joint weights $\tau_{j}$ that scale $\mathcal{L}_{acc}$.
IBP is a training-time regulariser that teaches the beat-stream network $\theta_b$ to produce latents that decode into physically smooth poses.

IBP penalises deviation from constant-velocity motion. The frozen decoder maps the predicted tokens to 6D joint rotations $\mathbf{x}_{j,t}$. The loss compares the current frame with the speed from the last two frames. The loss is the mean squared error against the constant-velocity prediction $\hat{\mathbf{x}}_{j,t} = 2\mathbf{x}_{j,t-1} - \mathbf{x}_{j,t-2}$, weighted by $\tau_{j,t}$:
$ \mathcal{L}_{acc} = \mathbb{E}_{j,t}\!\left[\, \tau_{j,t}\, \bigl\| \mathbf{x}_{j,t} - \hat{\mathbf{x}}_{j,t} \bigr\|^{2} \right]$.
Rather than penalising all joints equally, IBP weights each joint by a smoothing coefficient derived from De Leva's anthropometric masses~\cite{de1996adjustments} and the S-VIB gate:
\begin{equation}
    \tau_{j,t} = \tau_{\mathrm{base}} \cdot \sqrt{\frac{m_j}{m_{\max}}} \cdot (1-\Psi_t) \cdot (1+\alpha\,\sigma^{2}_{\phi,t}),
    \label{eq:tau}
\end{equation}

where $m_j$ is the De Leva body-segment mass fraction for joint $j$, $m_{\max}$ is the fraction of the heaviest segment (spine1/abdomen, ${\approx}0.163$ of body mass), so $\sqrt{m_j/m_{\max}}\in(0,1]$ is a sqrt-compressed relative inertial weight; $\tau_{\mathrm{base}}\!=\!0.5$ is the maximum smoothing applied to the heaviest arm-chain joint on a pure beat frame.
Hence, a uniform IBP gives each joint the same weight. De Leva gives more weight to heavier body parts and less to the hands. The gate lowers the loss on meaning frames.

Hence, training with this regulariser encourages smoother latents without any explicit physics at inference time. The face latent $Z_e^{\mathrm{b}}$ does not pass through the IBP block since facial motion has distinct dynamics that should not be constrained by IBP.

\textbf{Semantic stream} (bottom, blue in Fig.~\ref{fig:hierarchical_refiner}) operates on MGSC output $\mathbf{S}^m$. We first add masked seed pose embeddings ($\tilde{\mathbf{p}}$) and $\mathrm{ID}$, along with periodic positional encodings, before a self-attention pass. The resulting embeddings are then integrated with  $\mathbf{S}^m$ through cross-attention. Then three region-specific MLPs project these into semantic region latents $Z_u^{\mathrm{s}}$, $Z_h^{\mathrm{s}}$, $Z_l^{\mathrm{s}}$.

Inspired by~\cite{zhang2025SemTalk}, a \emph{Hierarchical Cross-Attention (HCA)} block refines these latents by allowing each region to attend to its sibling semantic latents: the hands decoder attends to $Z_u^{\mathrm{s}} + Z_l^{\mathrm{s}}$ as \texttt{[K,V]}, 
the upper decoder attends to $Z_h^{\mathrm{s}} + Z_l^{\mathrm{s}}$, and the lower decoder attends to $Z_u^{\mathrm{s}} + Z_h^{\mathrm{s}}$, producing final semantic latents $\hat{Z}_u^{\mathrm{s}}$, $\hat{Z}_h^{\mathrm{s}}$, $\hat{Z}_l^{\mathrm{s}}$.  HCA is applied similarly on the beat codebooks. Finally, as illustrated in
Fig.~\ref{fig:pipeline}(c), for each of the three body regions
$r\!\in\!\{h,u,l\}$, the beat residual $Z_r^{\mathrm{b}}$
(yellow token, $\Psi\!\approx\!0$)
and the semantic residual $Z_r^{\mathrm{s}}$ (blue token, $\Psi\!\approx\!1$) are blended
frame-by-frame via a fusion function.
The fused and face latents are then quantised by nearest-neighbour lookup in the RVQ codebook:

\[
\begin{aligned}
Z_r
&=
(1-\Psi)\hat{Z}^{\mathrm{b}}_{r}
+\Psi\hat{Z}^{\mathrm{s}}_{r}, \\
\hat{c}_{r,t}
&=\arg\min_{k\in\{1,\dots,K\}}
\left\|e_k-Z_{r,t}\right\|_2^2 .
\end{aligned}
\]
where \(e_k\) denotes the \(k\)-th codebook vector and \(\hat{c}_{r,t}\) is the selected discrete token for region \(r\) at frame \(t\).
The selected tokens are decoded by the frozen Stage-1 decoder as $
\hat{G}_r = D_r(\hat{c}_r)$.
Finally, our \textbf{full training objective} is:
\begin{equation}
    \mathcal{L} = \mathcal{L}_{\mathrm{lat}} + \mathcal{L}_{\mathrm{cls}} + \mathcal{L}_{\mathrm{sem}} + \beta_{\mathrm{vib}}\,\mathcal{L}_{\mathrm{kl}} + \beta_{\mathrm{phys}}\,\mathcal{L}_{\mathrm{acc}},
    \label{eq:total_loss}
\end{equation}
where: $\mathcal{L}_{\mathrm{lat}}$ is the MSE between the predicted continuous latents and the Stage-1 VQ targets for all body regions; $\mathcal{L}_{\mathrm{cls}}$ is the cross-entropy over the four RVQ codebook levels for each region; $\mathcal{L}_{\mathrm{sem}}$ (Sec.~\ref{sec:svib}) is the S-VIB gate supervised against BEAT2 semantic flags; $\mathcal{L}_{\mathrm{kl}}$ (Sec.~\ref{sec:svib}) is the VIB bottleneck KL with a free-bits floor; and $\mathcal{L}_{\mathrm{acc}}$ (Sec.~\ref{sec:ibp}) is the IBP inertia residual, active only on beat frames via $\tau_{j,t}$. The scalars $\beta_{\mathrm{vib}}$ and $\beta_{\mathrm{phys}}$ are warmup-scheduled weights.

\section{Experiments}
\paragraph{Dataset.}
We evaluate DuoGesture on BEAT2~\cite{liu2024emage}, a standard benchmark for co-speech gesture generation. It contains approximately 76 hours of speech, motion, facial expression, and speaker identity annotations from 30 speakers, covering diverse expressive behaviours. Unlike other datasets, such as TalkSHOW~\cite{yi2023generating}, BEAT2 provides frame-level annotations of gesture type (including semantic and beat gestures) and eight emotion categories, making it suitable for our study. Following the standard BEAT2 protocol used in prior holistic gesture-generation work~\cite{liu2024emage,mughal2025retrieving}, we report results in two settings: \emph{(i) a single-speaker setting on Speaker~2} and \emph{(ii) a multi-speaker setting over 25 speakers}. Unless otherwise stated, the data are split into training, validation, and test partitions at 85\%/7.5\%/7.5\%. The \textbf{implementation details} are elaborated in Sections 2 and 3 of the Supplementary materials.

\begin{table*}[th]
\centering
\caption{Comparison on BEAT2 (single speaker, top; all speakers, bottom).  Under the same protocol, rows marked $^{*}$ are taken from PyraMotion~\cite{yinpyramotion} and all other rows are our own runs. SRGR and relative LDLJ are therefore unavailable for $^{*}$ rows. Best, second, and third are marked per column within each block.}

\setlength{\tabcolsep}{4pt}
\resizebox{0.95\textwidth}{!}{
\begin{tabular}{c l c c c c c c c}
\toprule
\textbf{Setting} & \textbf{Model}
  & FGD $\times 10^{-1}$ $\downarrow$
  & BA $\times 10^{-1}$ $\uparrow$
  & Diversity $\uparrow$
  & MSE $\times 10^{-3}$ $\downarrow$
  & LVD $\times 10^{-5}$ $\downarrow$
  & SRGR $\uparrow$
  & Rel.\ LDLJ $\rightarrow 0$ \\
\midrule

\multirow{10}{*}{\rotatebox[origin=c]{90}{One Speaker}}
& TalkSHOW~\cite{yi2023generating} (CVPR 2023)
    & 6.209 & 6.947 & 13.47 & 7.791 & 7.771 & 0.248 & 1.6520 \\
& AMUSE\textsuperscript{*}~\cite{chhatre2024emotional} (CVPR 2024)
    & 12.11 & \bestcell{8.318} & \bestcell{14.93} & -- & -- & -- & -- \\
& ProbTalk\textsuperscript{*}~\cite{liu2024towards} (CVPR 2024)
    & 5.040 & 7.711 & 13.27 & 8.617 & -- & -- & -- \\
& EMAGE~\cite{liu2024emage} (CVPR 2024)
    & 5.512 & 7.724 & 13.06 & 7.680 & 7.556 & 0.314 & \secondcell{0.0470} \\
& MambaTalk\textsuperscript{*}~\cite{xu2024mambatalk} (NeurIPS 2024)
    & 5.366 & \secondcell{7.812} & \secondcell{13.95} & \bestcell{6.289} & \bestcell{6.897} & -- & -- \\
& RAG-Gesture\textsuperscript{*}~\cite{mughal2025retrieving} (CVPR 2025)
    & 8.080 & 7.340 & 11.97 & -- & -- & -- & -- \\
& GestureLSM~\cite{liu2025gesturelsm} (ICCV 2025)
    & \secondcell{4.247} & 7.290 & \thirdcell{13.76} & \thirdcell{7.139} & 7.469 & \secondcell{0.320} & 1.5138 \\ %
& SemTalk~\cite{zhang2025SemTalk} (ICCV 2025)
    & \thirdcell{4.278} & \thirdcell{7.770} & 12.91 & 7.153 & \secondcell{6.938} & \thirdcell{0.315} & \thirdcell{-0.5910} \\
& PyraMotion\textsuperscript{*}~\cite{yinpyramotion} (NeurIPS 2025)
    & 4.612 & 7.420 & 13.42 & 7.176 & \thirdcell{7.270} & -- & -- \\
& \textbf{Ours (DuoGesture)}
    & \bestcell{4.101} & 7.557 & 12.34 & \secondcell{7.103} & 7.646 & \bestcell{0.343} & \bestcell{-0.0304} \\
\midrule

\multirow{5}{*}{\rotatebox[origin=c]{90}{All Speakers}}
& TalkSHOW~\cite{yi2023generating} (CVPR 2023)
    & 6.145 & 6.863 & \bestcell{13.12} & 7.953 & 7.824 & 0.257 & 1.7360 \\
& EMAGE~\cite{liu2024emage} (CVPR 2024)
    & 5.643 & \bestcell{7.707} & \secondcell{12.92} & \thirdcell{7.694} & \secondcell{7.593} & \secondcell{0.326} & \secondcell{0.0530} \\
& GestureLSM~\cite{liu2025gesturelsm} (ICCV 2025)
    & \secondcell{4.268} & 5.250 & 11.20 & \secondcell{7.582} & 7.685 & \thirdcell{0.305} & 1.5386 \\
& SemTalk~\cite{zhang2025SemTalk} (ICCV 2025)
    & \thirdcell{5.214} & \thirdcell{7.689} & 12.74 & \thirdcell{7.612} & \bestcell{7.498} & 0.302 & \thirdcell{-0.6020} \\
& \textbf{Ours (DuoGesture)}
    & \bestcell{4.081} & \secondcell{7.699} & \thirdcell{12.83} & \bestcell{7.502} & \thirdcell{7.658} & \bestcell{0.348} & \bestcell{-0.0390} \\
\bottomrule
\end{tabular}}
\label{tab:main}
\end{table*}

\paragraph{Evaluation Metrics.}
We evaluate  co-speech gestures along several dimensions: distributional realism,
speech--motion synchrony, motion variation, and facial stability. We treat user studies and
Fr\'echet Gesture Distance (FGD)~\cite{yoon2020speech} as the primary evaluations, since FGD compares generated and real motion distributions in a learned
gesture-feature space, and is the standard metric with the strongest reported
perceptual support, where FGD was the only objective metric found to correlate with subjective human-likeness
ratings~\cite{kucherenko2024evaluating}.

We report two additional metrics alongside FGD since they test the contributions we make. \textbf{Semantic-Relevant Gesture Recall}
(SRGR)~\cite{liu2022beat} weights gesture recall of annotated semantic gestures. It scores motion only where a gesture is expected to be semantic, which is what MGSC (Sec.~\ref{sec:MGSC})
targets. \textbf{Relative log dimensionless jerk}
($\Delta\mathrm{LDLJ}$)~\cite{balasubramanian2015analysis} measures smoothness
as a signed difference from ground truth,
$\Delta\mathrm{LDLJ} = \mathrm{LDLJ}_{\mathrm{gen}} - \mathrm{LDLJ}_{\mathrm{GT}}$,
where positive values indicate over-smoothing and negative values indicate excess jitter. Relative LDLJ tests the effectiveness of IBP (Sec.~\ref{sec:ibp}).
In addition, we include Beat Alignment (BA)~\cite{li2021ai}, pairwise L1
Diversity~\cite{li2021audio2gestures}, facial MSE, and L1 Vertex Difference
(LVD)~\cite{xing2023codetalker} as secondary diagnostic metrics. These metrics
are informative but not sufficient indicators of perceptual quality. High BA can
reflect exaggerated beat-like motion, and high Diversity can reflect large
motion variance rather than plausible or semantically appropriate gestures.

\paragraph{Comparison Methods.}
We benchmark DuoGesture against a broad set of representative holistic
co-speech gesture generators, covering early speech-to-gesture models,
hierarchical and diffusion-based methods, semantic-based approaches,
and recent BEAT2 state-of-the-art systems. The comparison includes 
RAG-Gesture~\cite{mughal2025retrieving},
ProbTalk~\cite{liu2024towards},
MambaTalk~\cite{xu2024mambatalk}, TalkSHOW~\cite{yi2023generating},
EMAGE~\cite{liu2024emage}, SemTalk~\cite{zhang2025SemTalk},
GestureLSM~\cite{liu2025gesturelsm}, and
PyraMotion~\cite{yinpyramotion}.

\subsection{Overall Quantitative Comparison}
Table \ref{tab:main} shows the BEAT2 comparison with state-of-the-art methods. DuoGesture achieves the lowest FGD in both the single-speaker and all-speaker settings ($4.101$ and $4.081$, respectively), outperforming GestureLSM and SemTalk. It also leads on both contribution-specific metrics: SRGR and $|\Delta\mathrm{LDLJ}|$. This shows that gestures are semantically better recalled and closer to the dynamics of real motion smoothness than the presented baseline. On the secondary diagnostics, DuoGesture remains competitive. For BA and Diversity, it sits within a small margin of the best baselines. It also achieves the best MSE in the all-speaker setting. However, methods with higher BA or Diversity, such as AMUSE, achieve this with a worse FGD.
Since FGD is the objective metric most consistently linked to perceived human-likeness \cite{kucherenko2024evaluating}, we interpret this as a favourable overall trade-off.

\subsection{Ablation Studies}
\label{sec:ablation}

\begin{table}[th]
\centering
\caption{MGSC input ablation (all-speaker settings).}
\setlength{\tabcolsep}{8pt}
\resizebox{0.9\columnwidth}{!}{
\begin{tabular}{l c c}
\toprule
MGSC condition & FGD $\downarrow$ & SRGR $\uparrow$ \\
\midrule
Full MGSC (TMR + emotion)  & \textbf{4.081} & \textbf{0.348} \\
Without emotion            & 4.110 & 0.335 \\
CLIP instead of TMR        & 4.436 & 0.318 \\
CLIP, without emotion      & 4.332 & 0.305 \\
\bottomrule
\end{tabular}}
\label{tab:mgsc_encoder_ablation}
\end{table}

\paragraph{Motion-grounded semantic conditioning.}
Table ~\ref{tab:mgsc_encoder_ablation} tests how MGSC improves semantic gesture recall by varying only its inputs. Replacing TMR \cite{petrovich2023tmr} with CLIP \cite{radford2021learning} substantially lowers performance, raising FGD from 4.081 to 4.436 and lowering SRGR from 0.348 to 0.318, whereas removing the emotion embedding has a minimal impact (4.110, 0.335).
Performance in terms of SRGR degrades monotonically as motion-grounded information is removed. Interestingly, FGD improves when CLIP is used without emotion embedding. In general, motion grounding of the representation is the source of the performance gain.

\begin{table}[th]
\centering
\caption{IBP ablation (all-speaker settings).}
\setlength{\tabcolsep}{3pt}
\resizebox{0.8\columnwidth}{!}{
\begin{tabular}{l c c c c}
\toprule
IBP variant
& FGD $\downarrow$
& BA $\uparrow$
& Diversity $\uparrow$
& \shortstack{$\Delta$LDLJ\\$\to 0$} \\
\midrule
No IBP        & 4.178 & 7.446 & 12.77 & $-0.4904$ \\
Uniform IBP   & 4.278 & 7.450 & 12.80 & $\phantom{-}1.2790$ \\
De Leva IBP   & \textbf{4.081} & \textbf{7.699} & \textbf{12.83} & $\mathbf{-0.0390}$ \\
\bottomrule
\end{tabular}}
\label{tab:ibp_ablation}
\end{table}

\paragraph{Inertial beat prior.}
The benefit of IBP could come from smoothing alone or specifically from the anthropometric weighting in Eq.~\ref{eq:tau}. Table~\ref{tab:ibp_ablation} separates the two by comparing no prior, a uniform variant applying the same $\tau$ to every arm-chain joint, and the full De Leva weighting.
Uniform smoothing is less effective, raising FGD from $4.178$ to $4.278$, worse than applying no prior at all. Most importantly, without IBP, $\Delta\mathrm{LDLJ}$ is ($-0.490$). Uniform smoothing results in over-smoothing with  $\Delta\mathrm{LDLJ}$ of ($+1.279$). However, the anthropometric weighting gives the best results, near ground-truth smooth motion ($-0.039$), while also producing the largest beat-alignment gain ($7.699$).

\begin{table}[th]
  \centering
  \caption{
    Component-wise ablation of DuoGesture (all-speakers).
  }
  \label{tab:ablation}
  \small
  \setlength{\tabcolsep}{3pt}
  \renewcommand{\arraystretch}{1.12}
  \resizebox{\linewidth}{!}{%
  \begin{tabular}{l ccc ccc}
  \toprule
  \textbf{Variant}
    & \textbf{MGSC}
    & \textbf{S-VIB}
    & \textbf{IBP}
    & \textbf{FGD} $\downarrow$
    & \textbf{BA} $\uparrow$
    & \textbf{Diversity} $\uparrow$ \\
  \midrule
  (a) w/o MGSC (S-VIB + IBP only)
    & -- & $\checkmark$ & $\checkmark$
    & 4.803 & 7.531 & 12.61 \\
  (b) MGSC only (linear $\sigma$-gate)
    & $\checkmark$ & -- & --
    & 4.306 & 7.551 & 12.52 \\
  (c) MGSC + S-VIB (no IBP)
    & $\checkmark$ & $\checkmark$ & --
    & 4.178 & 7.446 & 12.77 \\
  (d) MGSC + IBP (linear $\sigma$-gate)
    & $\checkmark$ & -- & $\checkmark$
    & 4.137 & 7.557 & 12.65 \\
  (e) Full DuoGesture
    & $\checkmark$ & $\checkmark$ & $\checkmark$
    & \textbf{4.081}
    & \textbf{7.699}
    & \textbf{12.83} \\
  \bottomrule
  \end{tabular}%
  }
\end{table}

\paragraph{Component-wise ablation.}
Table \ref{tab:ablation} checks the contribution of each main module in the all-speaker setting: each row removes or replaces one component while holding the others fixed. 
MGSC dominates the FGD gain. This involves removing all three MGSC components: emotion embeddings and timing from FastText — and replacing motion-grounded semantic representations with standard semantic representations. This raises the FGD from 4.081 to 4.803. The effect of this is larger than that of removing S-VIB (+0.056) or IBP (+0.097).  
This is consistent with our hypothesis that motion-grounded semantic conditioning closes the linguistic–kinematic gap left open by purely lexical embeddings. 

IBP improves beat alignment, conditional on the stochastic gate. With S-VIB, adding IBP raises BA from 7.446 to 7.699, the largest BA change in the table. Under the deterministic gate, BA changes minimally ($7.551$ to $7.557$). The two components are therefore complementary, mirroring Table \ref{tab:ibp_ablation}.
Replacing S-VIB with the deterministic $\sigma$ gate (variants b and d) lowers diversity to 12.52 and 12.65 versus 12.83 for the full model. 

Hence, the ablation supports the design claim that semantic grounding, stochastic stream selection, and biomechanical beat regularisation are useful individually and most effective when combined.

\subsection{Subjective and Qualitative Results}
\subsubsection{User study.}
We conduct a perceptual study using 35-second clips from the BEAT2
test set. The study includes six narrated topics and compares Ground Truth and representative state-of-the-art methods, namely,
EMAGE, SemTalk, and DuoGesture. Thirty native English-speaking participants
from the UK, balanced by self-reported gender (male: female = 1:1; mean age
38.6), evaluated 24 randomly ordered videos using a five-point Likert scale.
Participants rated each video along three axes: naturalness, motion diversity,
and alignment with speech content and timing. The order of methods and clips was randomised for each participant.

\par
\subsubsection{User Studies Interface and Details}
\begin{figure}[t]
  \centering
  \includegraphics[width=1\linewidth]{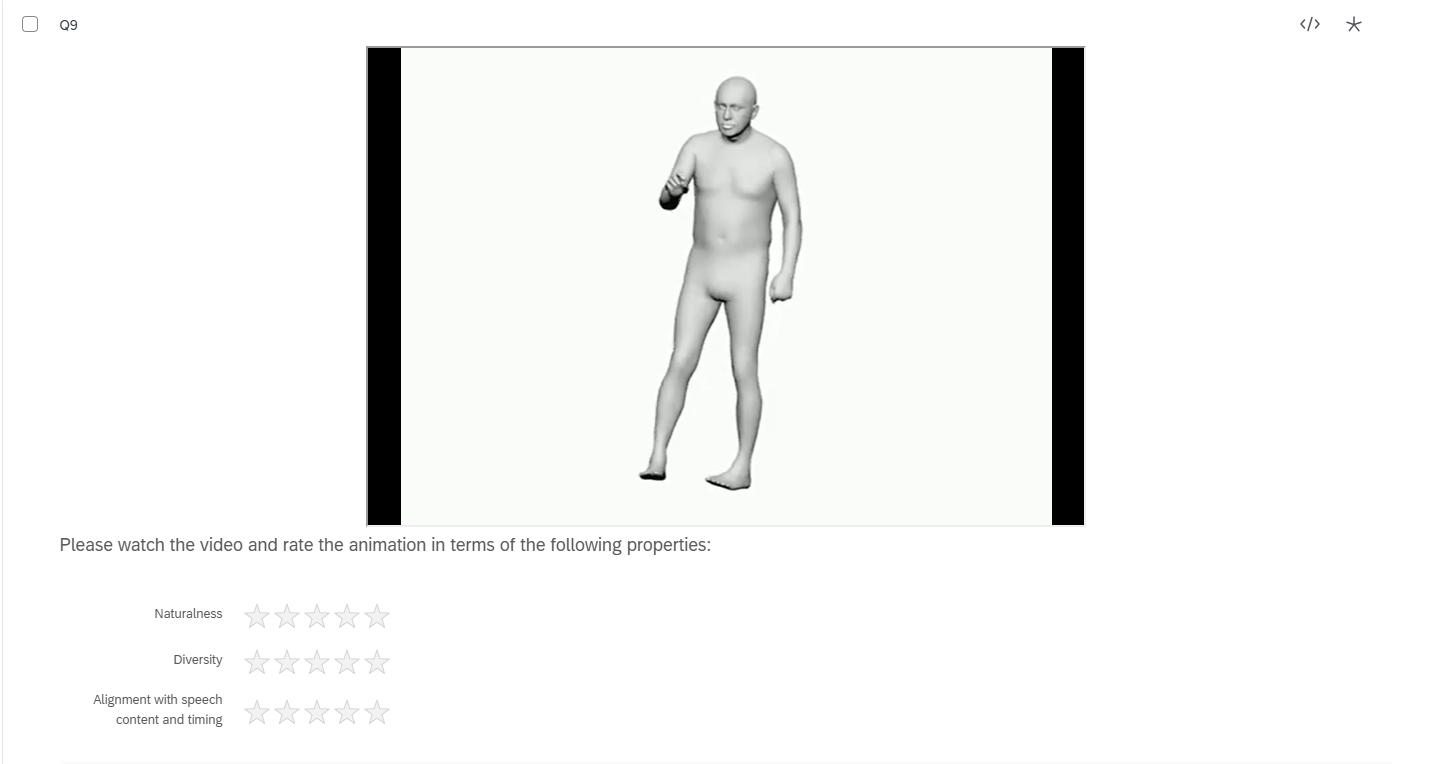}
  \caption{User study interface developed using Qualtrics. Participants were instructed to watch a co-speech gesture animation video and to evaluate the generated motions using a five-point Likert scale across three aspects: naturalness, diversity, and alignment with speech content and timing.}
  \label{fig:u1}
\end{figure}

Figure~\ref{fig:u1} presents the evaluation interface used in our user study, which was developed in Qualtrics. During the evaluation, participants were required to first watch a co-speech gesture animation video and subsequently assess the quality of the generated gestures using a five-point Likert scale. The evaluation focused on three aspects: \emph{Naturalness}, \emph{Diversity}, and \emph{Alignment with Speech Content and Timing}. In total, the questionnaire consisted of 24 video samples for evaluation. To minimise potential ordering effects and subjective bias, all videos were displayed in a randomised order for each participant. Participant recruitment and study administration were conducted through the Prolific platform, targeting participants from English-speaking countries.

\begin{figure}[t]
  \centering
  \includegraphics[width=1\linewidth]{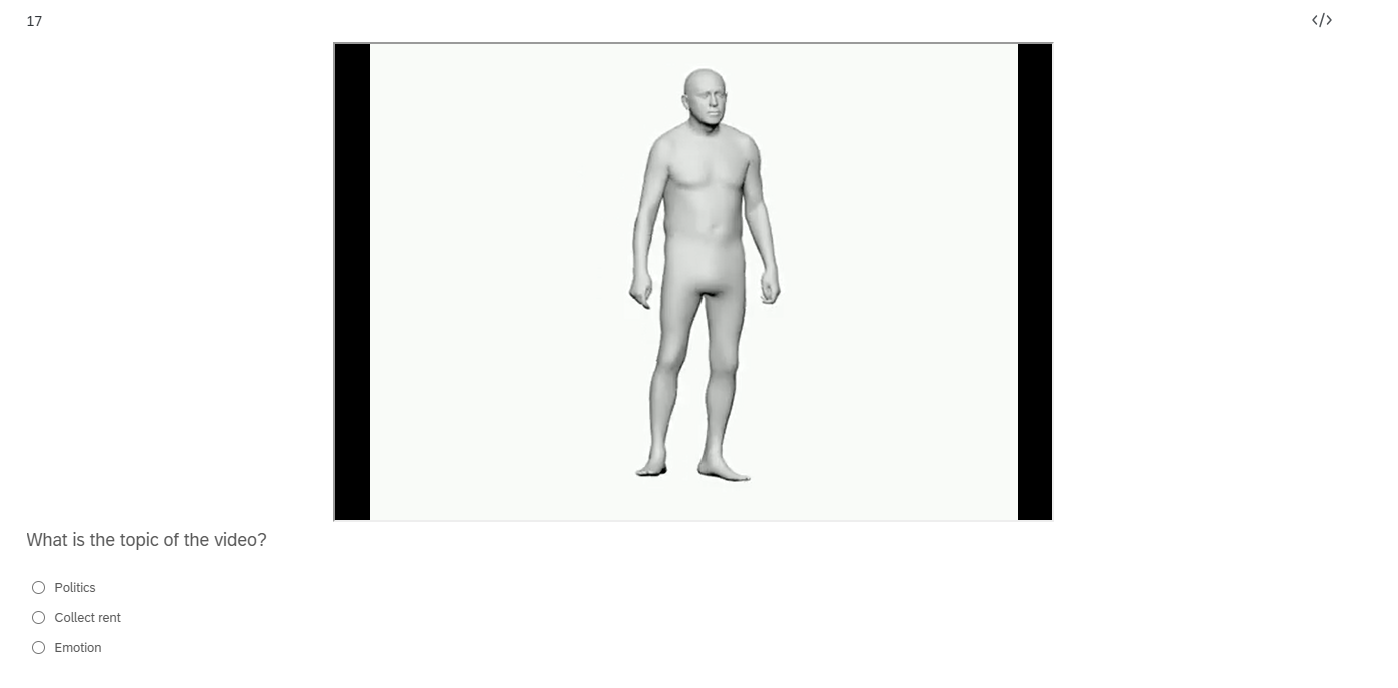}
  \caption{Attention-check interface employed during the user study to verify participant engagement and comprehension of the narrated content.}
  \label{fig:u2}
\end{figure}

To further ensure the reliability and validity of the collected responses, we incorporated additional attention-check questions throughout the evaluation process, as illustrated in Figure~\ref{fig:u2}. In these questions, participants were asked to identify the topic of the narrated speech from several candidate options. Responses from participants who failed the attention-check questions were excluded from the final analysis. These quality-control measures were introduced to verify that participants carefully watched the videos and adequately understood the accompanying speech content during the evaluation.

\begin{figure*}[ht]
  \centering
  \includegraphics[width=0.9\linewidth]{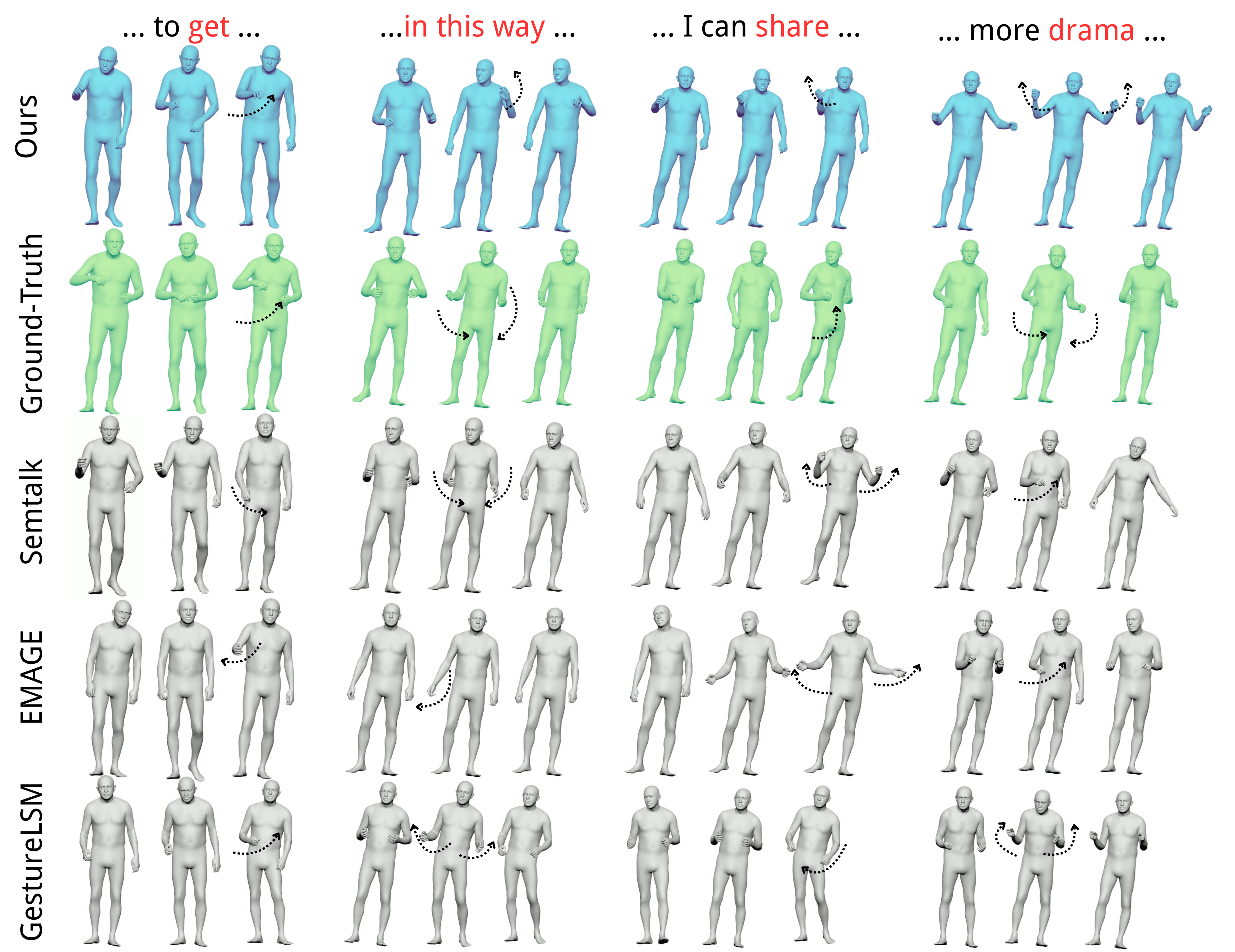}
  \caption{Qualitative comparison of co-speech gesture generation across
  semantic and beat-dominant speech contexts. We visualise motion sequences
  conditioned on representative phrases, including ``to get'', ``in this way'',
  ``I can share'', and ``more drama''.}
  \label{fig:qualitative}
\end{figure*}

\subsubsection{Results.} 
\begin{figure}[t]
  \centering
  \includegraphics[width=0.999\linewidth]{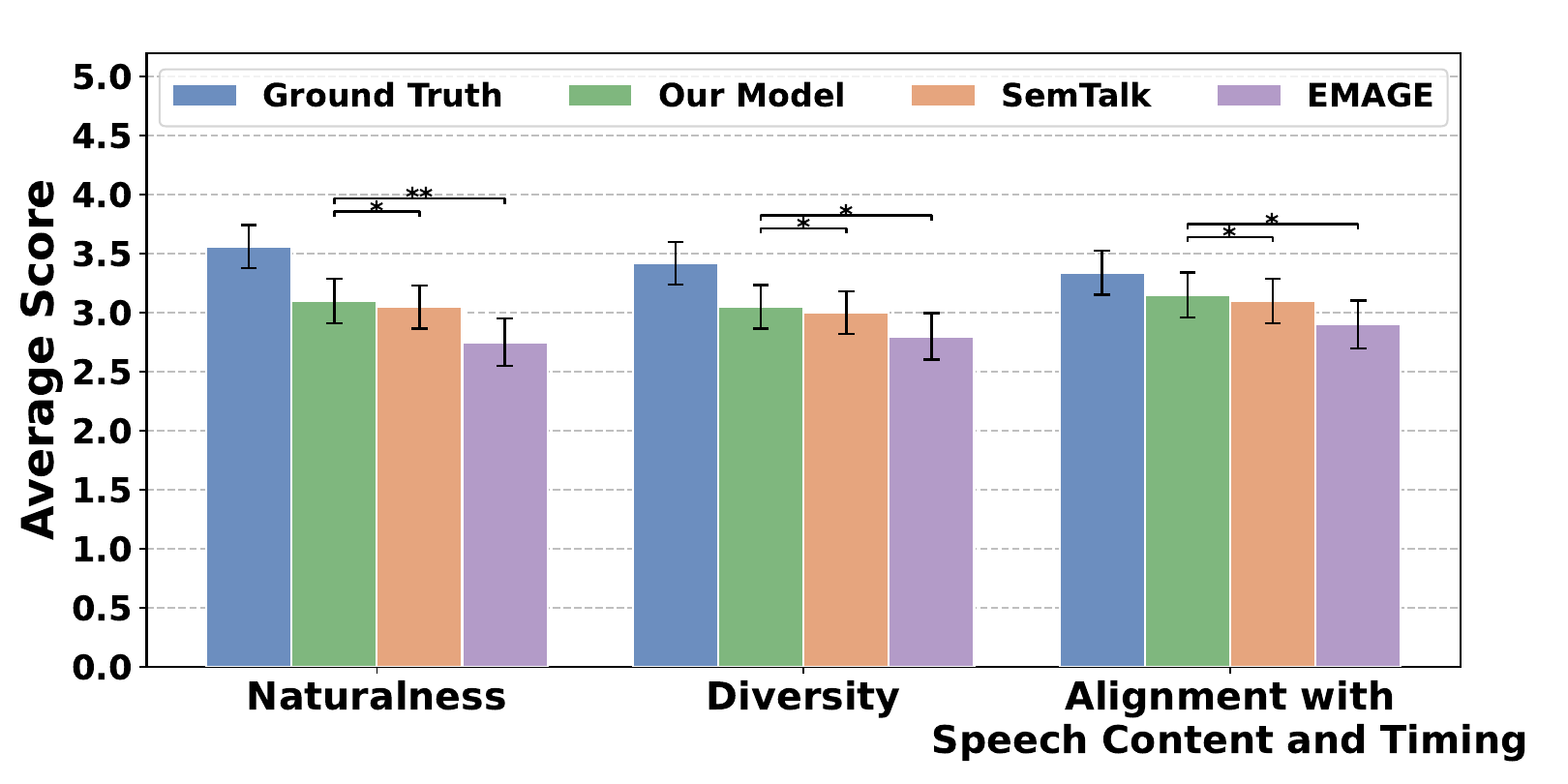}
  \caption{
    User study results comparing Ground Truth, DuoGesture,
    SemTalk, and EMAGE. Stars show significant differences.
  }
  \label{fig:userstudy}
\end{figure}

As shown in Fig. \ref{fig:userstudy}, Ground Truth receives the highest scores,
as expected. Among the generated visualisations, DuoGesture obtains the strongest overall
perceptual ratings, with higher judgments than SemTalk and EMAGE across the evaluated axes. This result is consistent with the quantitative analysis:
DuoGesture gives a better perceptual trade-off between realism, synchrony, and expressiveness. The user
study, therefore, provides independent evidence that the observed results in Table~\ref{tab:main} correspond to motions that users judge
as more natural and better aligned with speech. 

\subsection{Qualitative Results}
\paragraph{Qualitative comparison.}
Figure \ref{fig:qualitative} illustrates a comparison of the generated motion sequences for speech segments containing semantic and beat phrases. For semantic phrases such as 'to get', 'I can share', and 'more drama', DuoGesture produces gestures with clearer arm trajectories and a more visible, phrase-dependent structure than the other methods. SemTalk and GestureLSM tend to generate weaker or more ambiguous movements, whereas EMAGE often produces more plausible, albeit less semantically differentiated, gestures. For the beat-dominant phrase 'in this way', the generated motion remains temporally coherent with the speech rhythm, yet avoids excessive smoothing.

\section{Conclusion}
\label{sec:conclusion}
We presented DuoGesture, a co-speech gesture generation approach that uses a semantic stream to generate lexically grounded gestures and a beat stream to generate rhythmic motion aligned to prosody. A stochastic frame-level gate coordinates the interaction between these streams. DuoGesture achieves the lowest FGD on BEAT2 in both the single-speaker and all-speaker settings by combining motion-grounded semantic conditioning, stochastic stream selection, and an inertial beat prior, while remaining competitive on alignment, diversity, and facial-motion fidelity. It also outperforms other approaches on SRGR and $|\Delta\mathrm{LDLJ}|$ in both settings, the two metrics that test our two main contributions: MGSC and IBP. Ablation studies and user testing confirm that these components provide complementary benefits, supporting the idea of an explicit design process for co-speech gesture generation.

\small
\bibliographystyle{ieee}
\bibliography{egbib}

\normalsize   %

\twocolumn[{\centering\Large\bf Supplementary Material\par\vspace{1.5em}}]

\setcounter{section}{0}\renewcommand{\thesection}{S\arabic{section}}
\setcounter{figure}{0}\renewcommand{\thefigure}{S\arabic{figure}}
\setcounter{table}{0}\renewcommand{\thetable}{S\arabic{table}}
\setcounter{equation}{0}\renewcommand{\theequation}{S\arabic{equation}}
\def\theHsection{S\arabic{section}}
\def\theHfigure{S\arabic{figure}}
\def\theHtable{S\arabic{table}}
\def\theHequation{S\arabic{equation}}

\section{Motion Analysis: Beat vs.\ Semantic Motion on BEAT2}
\label{sec:gt_analysis}

\paragraph{Setup and controlled sampling.}
We analyse BEAT2~\cite{liu2022beat, liu2024emage} test-split motion ($\geq\!15$-frame windows, all 25 speakers) using the dataset's per-frame semantic annotation flags to segment sequences into contiguous \emph{beat windows} (flag $= 0$) and \emph{semantic windows} (flag $= 1$). To ensure a fair, balanced comparison, we draw a matched sample: for each speaker we randomly draw the same number of beat and semantic windows (capped at the smaller set), yielding $\approx\!1{,}870$ windows per class from 265 clips. All metrics are duration-weighted and accompanied by 95\% CIs from hierarchical bootstrap (speaker $\to$ clip $\to$ window). The matched windows are representative of typical gesture production for each class (short rhythmic repetitions for beat, and clear lexical stroke events for semantic), and the class balance removes any confound from the natural 12\% semantic sparsity.

\begin{figure*}[h]
    \centering
    \includegraphics[width=\linewidth]{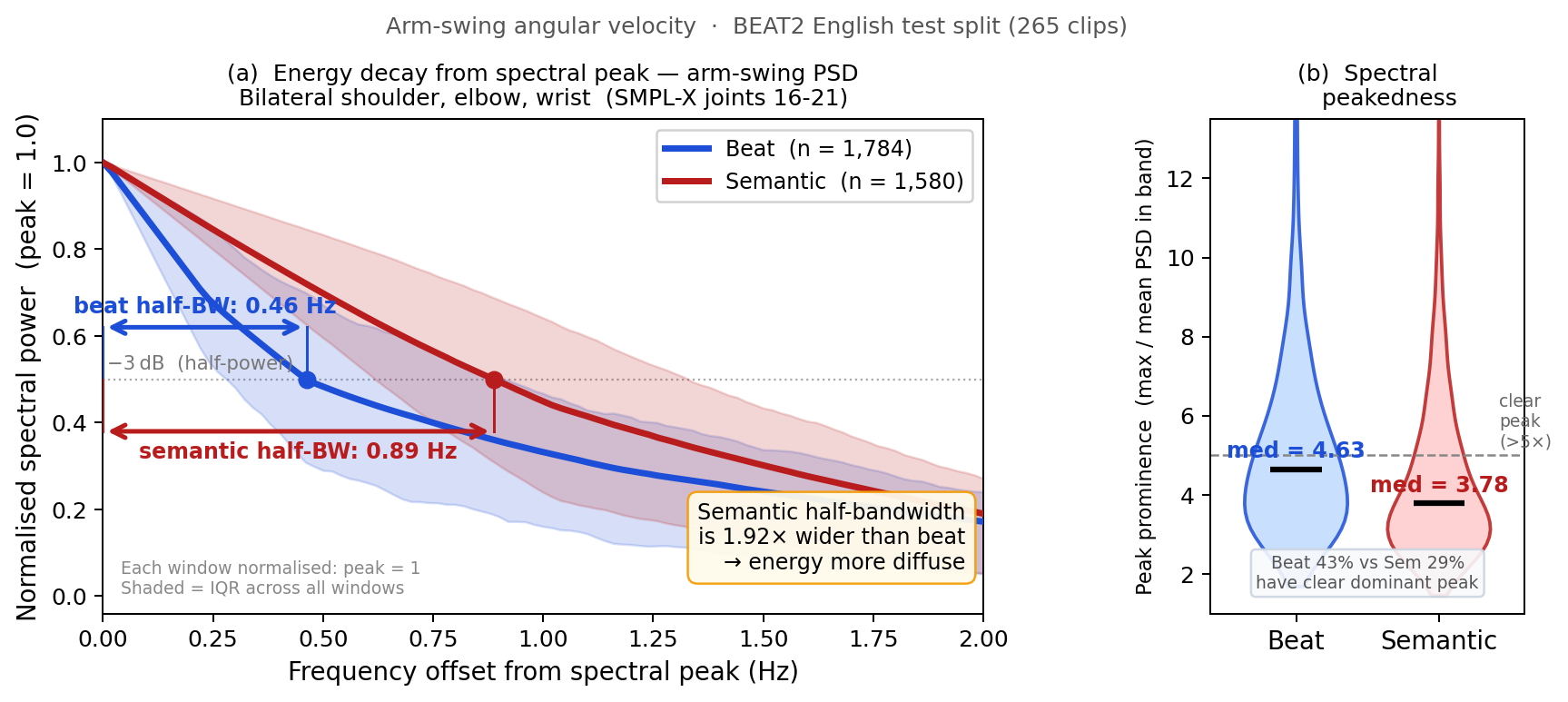}
    \caption{Arm-swing spectral analysis for bilateral shoulder/elbow/wrist joints (SMPL-X joints 16--21), matched beat and semantic windows (BEAT2 test split, $\approx\!1{,}870$ windows each, 25 speakers). \textbf{(a)}~Mean normalised PSD decay from the spectral peak (shaded = IQR): beat spectra (blue) have a half-bandwidth of $0.46$ Hz; semantic spectra (red) are $1.9\times$ broader ($0.89$ Hz), indicating diffuse rather than tonal energy. \textbf{(b)}~Peak prominence (max/mean PSD in band): 43\% of beat windows have a clear dominant peak ($>5\times$) versus only 29\% of semantic windows, confirming that rhythmic regularity is a beat-specific property.}
    \label{fig:arm_swing_psd}
\end{figure*}

\paragraph{Results.}
The two classes show clearly distinct kinematics (Table~\ref{tab:gt_analysis} and Fig.~\ref{fig:arm_swing_psd}). Beat motion peaks at $1.12$ Hz with moderate inter-joint coupling (PLV $= 0.31$), indicating that shoulder and forearm oscillate at partially independent rhythms,  consistent with a damped-pendulum model. A constant-velocity (inertial) oscillator explains $R^2 = 0.41$ of beat variance, and this improves by $\Delta R^2 = +0.84$ when the mass grouping covers arm $+$ core joints, but \emph{degrades} by $\Delta R^2 = -0.95$ when all 55 joints are included,  directly motivating IBP's arm-chain-only mask. Semantic motion peaks at $1.69$ Hz with strongly higher inter-joint coupling (PLV $= 0.53$, $\Delta\mathrm{PLV} = +0.22$), indicating holistic co-activation of the arm during a lexical stroke,  a pattern better captured by shape-conditioned generation than by a mass-weighted smoother. Arm-swing PSD (Fig.~\ref{fig:arm_swing_psd}) further shows that beat spectra have a narrow, peaked structure (half-BW $0.46$ Hz, prominence $4.6\times$) while semantic spectra are $1.9\times$ broader and less tonal, confirming that beat motion is rhythmically regular and semantic motion is not.

\begin{table*}[h]
\centering
\caption{Kinematic statistics for matched beat and semantic windows (BEAT2 test split, 25 speakers; hierarchical bootstrap 95\% CI in brackets).}
\label{tab:gt_analysis}
\setlength{\tabcolsep}{7pt}
\begin{tabular}{lcc}
\toprule
\textbf{Metric} & \textbf{Beat} & \textbf{Semantic} \\
\midrule
Shoulder peak (Hz) & 1.12\ [1.02, 1.23] & 1.69\ [1.56, 1.86] \\
PLV (shoulder $\leftrightarrow$ forearm) & 0.308\ [0.268, 0.349] & 0.525\ [0.487, 0.557] \\
Oscillator $R^2$ & 0.406\ [0.388, 0.427] & 0.509\ [0.493, 0.524] \\
$\Delta R^2$: arm+core vs.\ shoulder & $+0.84$ & --- \\
$\Delta R^2$: all joints vs.\ shoulder & $-0.95$ & --- \\
Arm-swing half-BW (Hz) & 0.46 & 0.89 \\
\bottomrule
\end{tabular}
\end{table*}

\section{Technical Implementation}
\label{appendix:implementation_details}

\paragraph{Data.} BEAT2~\cite{liu2024emage}, 85\%/7.5\%/7.5\% split, 64-frame clips at 30 fps (stride 20), 4-frame seed pose.

\paragraph{Training.} Adam ($\text{lr}=10^{-4}$, no weight decay), step LR decay ($\gamma=0.3$), 200 epochs. Stage~2 trained on 4 GPUs A100 via PyTorch DDP, effective batch size 256.

\paragraph{Representations.} Motion: 55 SMPL-X joints in rot6d (330-dim). Audio: HuBERT (1024-dim). Text: FastText~\cite{bojanowski2017enriching} (300-dim). All streams projected to 256-dim.

\paragraph{Stage~1.} Regional RVQ-VAE~\cite{liu2024emage}, codebook $C=256$, 4 RVQ levels, frozen during Stage~2.

\paragraph{Stage~2.} Transformer backbone: hidden size 768, 1 self-attention layer; 3 decoder layers and 1 codebook head per region. MGSC embeddings ($e_m$, $e_\varepsilon$) from TMR~\cite{petrovich2023tmr} (256-dim). S-VIB: bottleneck $Z=16$, timing projection 64-dim, KL warmup epochs 20--100 ($\beta_{\mathrm{target}}=0.01$), free-bits $\lambda_{\mathrm{fb}}=0.5$ nats, semantic boost $\rho=3.0$. IBP: physics warmup epochs 30--80 ($\lambda=0.01$), $\tau_{\mathrm{base}}=0.5$, $\alpha=1.0$, arm-chain mask only.

\section{Architecture Details}
\label{appendix:architecture}

Table~\ref{tab:arch} summarises the key dimensions of each Stage-2 component.

\begin{table*}[h]
\centering
\caption{Stage-2 architecture summary. All transformer layers use $H\!=\!4$ heads, dropout 0.1, and batch-first layout. Stage-1 weights are frozen.}
\label{tab:arch}
\setlength{\tabcolsep}{6pt}
\renewcommand{\arraystretch}{1.1}
\begin{tabular}{l l l}
\toprule
\textbf{Component} & \textbf{Details} & \textbf{Dims} \\
\midrule
\multicolumn{3}{l}{\textit{Inputs}} \\
HuBERT encoder & 2× Conv1d (k=3, p=1), BN, GELU & $1024 \to 256$ \\
Onset/amplitude & 2-layer MLP & $3 \to 256$ \\
FastText & Linear & $300 \to 256$ \\
Seed pose & VQEncoderV6 (3 layers, frozen) & $337 \to 768$ \\
Speaker ID & Embedding table (25 entries) & $\to 768$ \\
\midrule
\multicolumn{3}{l}{\textit{Shared backbone (both streams)}} \\
Self-attention & TransformerEncoder, 1 layer, $d_{\mathrm{ff}}$=1536 & $d=768$ \\
Audio cross-attention & TransformerDecoder, 8 layers, $d_{\mathrm{ff}}$=1536 & $d=768$ \\
\midrule
\multicolumn{3}{l}{\textit{MGSC}} \\
TMR projections~\cite{petrovich2023tmr} & Linear ($e_m$, $e_\varepsilon$) & $256 \to 256$ \\
Softmax memory gate & 2× Linear & $256 \to 1$ \\
Semantic cross-attention & TransformerDecoder, 1 layer, $d_{\mathrm{ff}}$=512 & $d=256$ \\
Output MLP & 2-layer & $256 \to 256$ \\
\midrule
\multicolumn{3}{l}{\textit{S-VIB}} \\
Timing projection & Linear + LN + GELU & $256 \to 64$ \\
Bottleneck heads & 2× Linear (mean, log-var) & $320 \to 16$ \\
Latent Sparsity Interpreter & 2-layer MLP + softmax & $16 \to 2$ \\
\midrule
\multicolumn{3}{l}{\textit{Region decoders (beat \& semantic, per region $r\in\{h,u,l\}$)}} \\
Region MLP & 2-layer & $768 \to 256$ \\
Temporal downsample & Conv1d, k=2, s=2 & $L \to L/2$ \\
HCA (3 decoders) & TransformerDecoder, 1 layer, $d_{\mathrm{ff}}$=512 & $d=256$ \\
\midrule
\multicolumn{3}{l}{\textit{Face (beat only)}} \\
Face decoder & TransformerDecoder, 4 layers, $d_{\mathrm{ff}}$=1536 & $d=768$ \\
Downsample & Linear + 2× Conv1d (k=2,s=2) + MLP & $768 \to 256$ \\
\midrule
\multicolumn{3}{l}{\textit{Codebook prediction heads (per region, 5 levels)}} \\
Autoregressive decoder & TransformerDecoder, 3 layers, $H\!=\!8$, $d_{\mathrm{ff}}$=1024 & $d=256$ \\
Level classifier & 2-layer MLP & $256 \to 256$ \\
\midrule
\textbf{Total Stage-2 params} & & $\approx 122$M \\
\bottomrule
\end{tabular}
\end{table*}

\noindent\textbf{Fusion.} For $r\in\{h,u,l\}$: $Z_r = (1{-}\Psi)\hat{Z}_r^b + \Psi\hat{Z}_r^s$, quantised as $\hat{c}_r = \arg\min_k\|e_k - Z_r\|_2$. The face region uses $Z_e^b$ directly (no semantic stream). All regions decoded by frozen $\mathcal{D}^r$: $\hat{\mathbf{G}}^r = \mathcal{D}^r(\hat{c}_r)$.

\small

\end{document}